\documentclass{bmvc2k}
\usepackage{algorithm}
\usepackage{algorithmic}
\usepackage{amsmath}
\usepackage{amssymb}
\usepackage{enumitem}
\usepackage{booktabs} 
\usepackage{multirow} 
\usepackage{xcolor}

\usepackage{amsmath,amsfonts,bm}

\renewcommand\vec[1]{\ensuremath\boldsymbol{#1}}

\newcommand{\mB}{\mathbf{B}}

\newcommand{\vy}{\mathbf{y}}

\newcommand{\mA}{\mathbf{A}}

\newcommand{\vx}{\mathbf{x}}

\newcommand{\mbr}[1]{\mathbb{R}^{#1}}

\newcommand{\mU}{\textbf{U}}
\newcommand{\mV}{\textbf{V}}

\newcommand{\mS}{\vec{S}}

\newcommand{\mW}{\boldsymbol{W}}

\newcommand{\stkout}[1]{{\ifmmode\text{\sout{\ensuremath{#1}}}\else\sout{#1}\fi}}

\title{SRLoRA: Subspace Recomposition in Low-Rank Adaptation via Importance-Based Fusion and Reinitialization}

\addauthor{Haodong Yang}{haodong.yang@anu.edu.au}{1}
\addauthor{Lei Wang}{l.wang4@griffith.edu.au}{2, 4, *}
\addauthor{Md Zakir Hossain}{zakir.hossain1@curtin.edu.au}{3, 1, 4}

\addinstitution{
 Australian National University
}
\addinstitution{
 Griffith University
}
\addinstitution{
 Curtin University
}
\addinstitution{
 Data61/CSIRO
}

\runninghead{Haodong Yang, Lei Wang, and Md Zakir Hossain}{Research Report}

\def\eg{\emph{e.g}\bmvaOneDot}

\def\ie{\emph{i.e}\bmvaOneDot}
\def\etal{\emph{et al}\bmvaOneDot}

\begin{document}

\maketitle

\renewcommand{\thefootnote}{\fnsymbol{footnote}}
\footnotetext[1]{Corresponding author.}

\begin{abstract}
Low-Rank Adaptation (LoRA) is a widely adopted parameter-efficient fine-tuning (PEFT) method that injects two trainable low-rank matrices ($\mA$ and $\mB$) into frozen pretrained models.
While efficient, LoRA constrains updates to a fixed low-rank subspace ($\Delta \mW = \mB\mA$), which can limit representational capacity and hinder downstream performance.
We introduce Subspace Recomposition in Low-Rank Adaptation (SRLoRA) via importance-based fusion and reinitialization, a novel approach that enhances LoRA's expressiveness without compromising its lightweight structure. SRLoRA assigns importance scores to each LoRA pair (a column of $\mB$ and the corresponding row of $\mA$), and dynamically recomposes the subspace during training. Less important pairs are fused into the frozen backbone, freeing capacity to reinitialize new pairs along unused principal directions derived from the pretrained weight’s singular value decomposition. This mechanism enables continual subspace refreshment and richer adaptation over time, without increasing the number of trainable parameters.
We evaluate SRLoRA on both language and vision tasks, including the GLUE benchmark and various image classification datasets.
SRLoRA consistently achieves faster convergence and improved accuracy over standard LoRA, demonstrating its generality, efficiency, and potential for broader PEFT applications.
\end{abstract}

\section{Introduction}
\label{sec:intro}

Fine-tuning large pretrained models is key to state-of-the-art results in vision and language \cite{devlin2019bertpretrainingdeepbidirectional} \cite{simonyan2015deepconvolutionalnetworkslargescale}, but full fine-tuning updates all parameters, causing high computational and storage costs \cite{hu2022lora}. This limits its use in multi-task adaptation and resource-limited environments.

To address these challenges, parameter-efficient fine-tuning (PEFT) methods \cite{hu2023llmadapters} have emerged as a compelling alternative.
These approaches reduce the number of trainable parameters by introducing lightweight modules \cite{houlsby2019parameterefficienttransferlearningnlp}\cite{lester2021powerscaleparameterefficientprompt} or modifications to the model \cite{li2021prefixtuningoptimizingcontinuousprompts}\cite{zaken2022bitfitsimpleparameterefficientfinetuning}, enabling efficient adaptation while retaining most of the pre-trained weights.
Among them, Low-Rank Adaptation (LoRA) \cite{hu2022lora} stands out for its simplicity and strong empirical performance.
LoRA introduces two low-rank trainable matrices, $\mA \in \mathbb{R}^{r \times n}$ and $\mB \in \mathbb{R}^{m \times r}$, to produce an update $\Delta \mW = \mB\mA$, while keeping the original weights $\mW \in \mbr{m \times n}$ frozen.
This allows the model to be fine-tuned by modifying only a small number of parameters.
%

Despite its efficiency, LoRA imposes a critical limitation: the parameter update is constrained to a fixed low-dimensional subspace.
This restriction can lead to suboptimal expressiveness and degraded performance, particularly in tasks that demand richer adaptation capacity \cite{hu2022lora}.
Once the low-rank directions are initialized, LoRA lacks a mechanism to explore or expand its update space during training.

\begin{figure}[tbp]
\centering
\begin{tabular}{cc}
\bmvaHangBox{\fbox{\includegraphics[height=5cm]{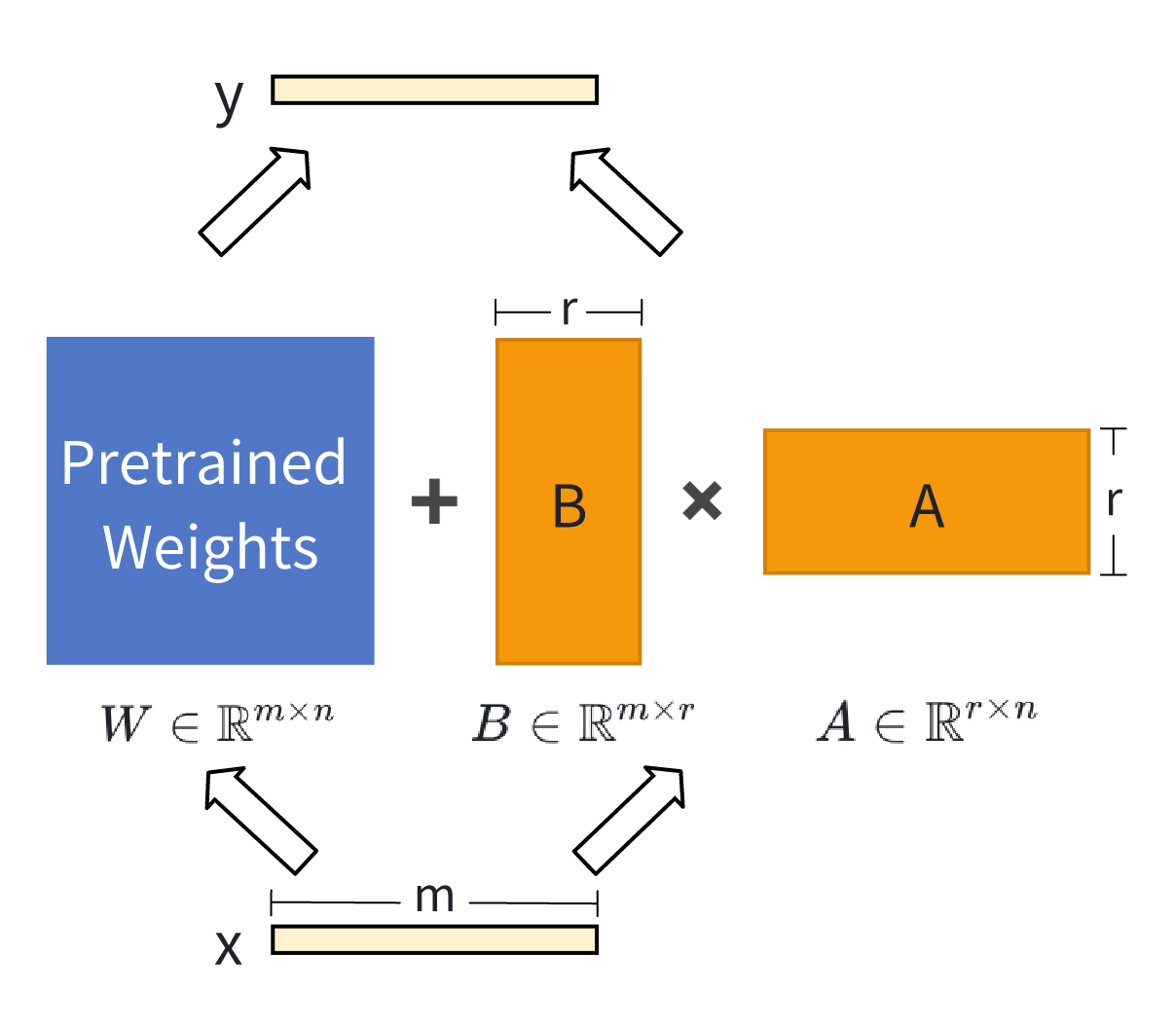}}} &
\bmvaHangBox{\fbox{\includegraphics[height=5cm]{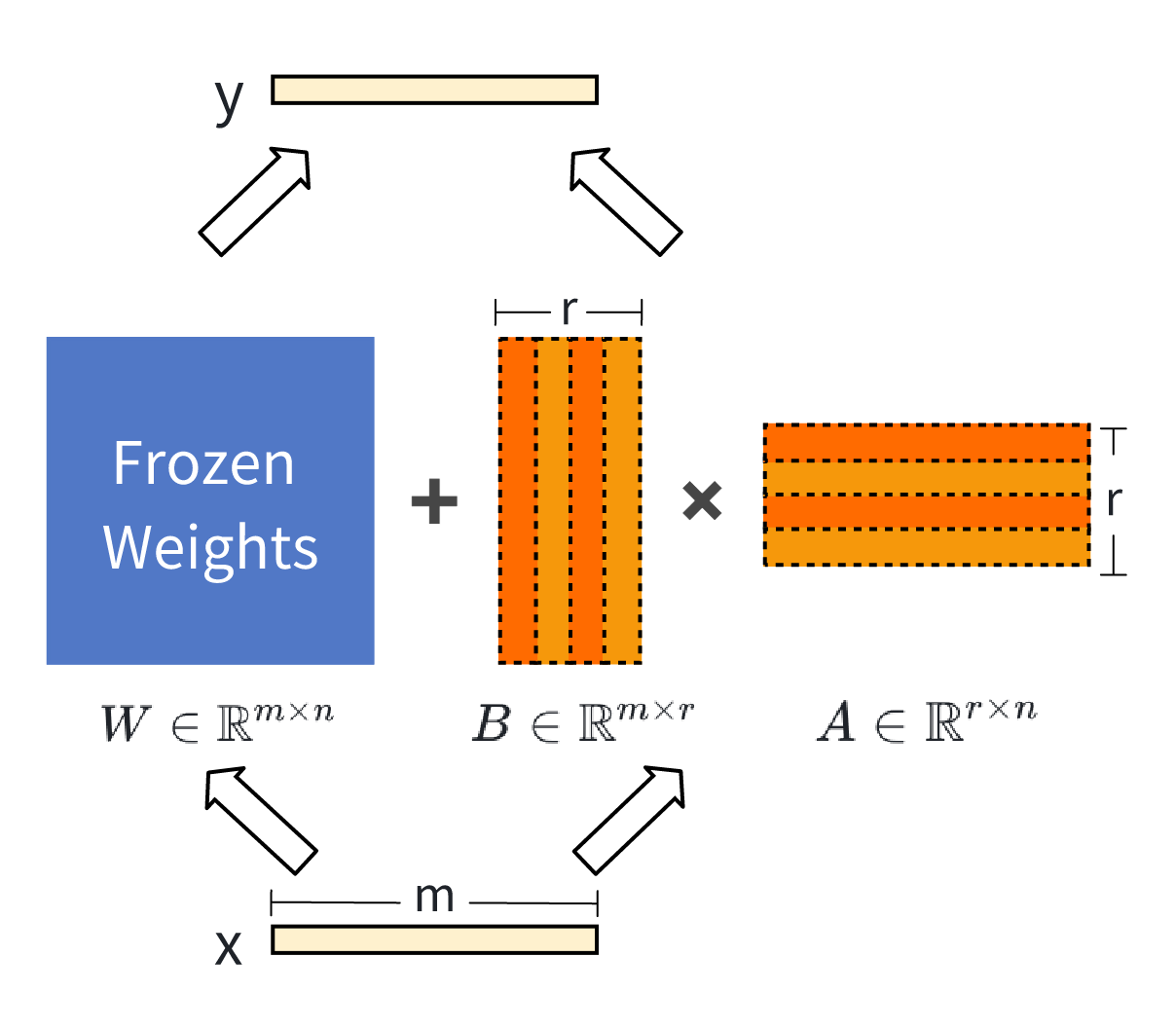}}} \\
\end{tabular}
\caption{Comparison with LoRA and SRLoRA. The blue areas represent pretrained/frozen parameters during training. Left: LoRA. The orange areas represent trainable parameters; Right: SRLoRA. Darker orange areas represent higher importance scores, and lighter orange areas represent lower importance scores.}
\label{fig:teaser}
\end{figure}

To overcome this limitation, we propose Subspace Recomposition in Low-Rank Adaptation (SRLoRA) via importance-based fusion and reinitialization. Our method enhances the flexibility of LoRA by dynamically modifying its update subspace throughout training, without changing LoRA's structural simplicity or increasing the number of trainable parameters (see Fig.~\ref{fig:teaser} for comparison).
The core idea is to identify less important update directions (\ie, pairs of columns in $\mB$ and corresponding rows in $\mA$), merge their contributions into the frozen pre-trained weights, and reinitialize them using unused principal directions derived from from the singular value decomposition (SVD) of the original weight matrix.
These reinitialized pairs are then adjusted to maintain consistency with the frozen backbone, enabling SRLoRA to recycle and explore new directions in the parameter space.


Through this subspace recomposition mechanism, SRLoRA significantly enhances the expressive capacity of LoRA while retaining its efficiency.
We evaluate SRLoRA across both vision and language tasks, including image classification with Vision Transformers and natural language understanding with DeBERTa-v3-base, demonstrating faster convergence and improved performance.
Our contributions are as follows:
\renewcommand{\labelenumi}{\roman{enumi}.}
\begin{enumerate}[leftmargin=0.6cm]
\item \textbf{We identify and address a core limitation of LoRA}: its inability to expand the update subspace beyond the initial low-rank directions, which restricts its fine-tuning capability.
\item \textbf{We propose SRLoRA}, a novel method that dynamically recomposes LoRA's update space by fusing less informative components into frozen weights and reinitializing them using unused SVD directions, without increasing the number of trainable parameters.
\item \textbf{We empirically validate SRLoRA} on both vision and language tasks, showing consistent improvements in convergence speed and final accuracy over standard LoRA, highlighting its generality and effectiveness.
\end{enumerate}


\section{Related Works}
\label{app:relatedwork}

\textbf{Parameter-Efficient Fine-Tuning (PEFT)} methods \cite{xu2023parameterefficientfinetuningmethodspretrained} have emerged as effective strategies to reduce the prohibitive computational and storage costs associated with full fine-tuning of large-scale pre-trained models \cite{han2024parameterefficientfinetuninglargemodels}. Instead of updating all model parameters, PEFT approaches introduce lightweight trainable components while keeping the majority of the model frozen, enabling scalable adaptation across diverse tasks.

\textbf{Low-Rank Adaptation.} Among PEFT methods, LoRA~\cite{hu2022lora} \cite{zhang2023adaptive}\cite{meng2024pissa} has gained significant popularity for its simplicity and strong performance. LoRA replaces full-rank weight updates with low-rank trainable matrices $\mA$ and $\mB$, producing an update $\Delta \mW = \mB\mA$. This results in the final output of the form: $\vy = (\mW + \mB\mA)\vx$,
where $\mW$ is the frozen pre-trained weight matrix. By significantly reducing the number of trainable parameters, LoRA enables efficient fine-tuning while achieving performance close to or on par with full fine-tuning.

%
However, the expressiveness of LoRA is fundamentally constrained by the fixed rank $r$ of the adaptation matrices. Once initialized, the update subspace remains static throughout training, potentially limiting its capacity to capture diverse task-specific variations.


\textbf{Adaptive rank methods.} To address the expressiveness bottleneck, several methods \cite{zhang2023adaptive}\cite{valipour2023dyloraparameterefficienttuning} have explored adaptive rank strategies. AdaLoRA~\cite{zhang2023adaptive} dynamically adjusts the rank during training based on a sensitivity-based scoring mechanism, allowing more effective allocation of parameter capacity across layers. DyLoRA~\cite{valipour2023dyloraparameterefficienttuning} trains LoRA modules across a range of ranks, identifying optimal configurations by analyzing representations at different ranks during training. These methods highlight the importance of subspace flexibility, demonstrating that a static low-rank approximation can be limiting for certain tasks.

\textbf{Initialization strategies.}
Another active research direction focuses on improving LoRA through smarter initialization of the low-rank matrices \cite{hu2022lora}\cite{meng2024pissa}\cite{wang2024loraga}. Poor initialization can hinder convergence and limit adaptation capacity (\eg, initialing $\mA$ with $\mathcal{N}(0, \sigma^2)$ and $\mB$ with $\mathbf{0}$). PiSSA~\cite{meng2024pissa} introduces a sensitivity-aware, SVD-based initialization that aligns the low-rank subspace with the most important directions of the pre-trained weights. Specifically, it initializes the LoRA matrices as:
$\mA = \mS_{[:r, :r]}^{1/2} \mV_{[:, :r]}^\top \in \mbr{r \times n}$, $\mB = \mU_{[:, :r]} \mS_{[:r, :r]}^{1/2} \in \mbr{m \times r}$,
where $\mU$, $\mS$, and $\mV$ are obtained from the SVD of the original weight matrix $\mW$. The residual component is represented as:
$\mathbf{W}^{\mathrm{res}} = \mU_{[:, r:]} \mS_{[r:, r:]} \mV_{[:, r:]}^\top$,
ensuring complementarity between the adapted and frozen weights.
Similarly, LoRA-GA~\cite{wang2024loraga} uses gradient-based information derived from singular vectors of $\mW$ to guide the initialization of $\mA$ and $\mB$, promoting alignment between the adaptation subspace and task-relevant gradients.

\textbf{Our perspective: dynamic subspace recomposition.} While prior work has focused on adapting the \textit{rank} or improving the \textit{initialization}, these strategies do not address the core limitation that LoRA’s update subspace remains fixed once initialized. In contrast, our method, \textbf{SRLoRA}, introduces a novel direction: \textit{dynamic subspace recomposition} within a fixed parameter budget.
Specifically, SRLoRA periodically identifies less informative adaptation directions (\ie, low-importance column-row pairs in $\mB$ and $\mA$), fuses them into the frozen pre-trained weights to preserve their contribution, and reinitializes the corresponding parameters using previously unused SVD components. This enables the model to \textit{recycle low-rank capacity}, allowing continuous exploration of new subspaces throughout training. Unlike adaptive-rank methods, our approach maintains a constant number of trainable parameters, striking a better balance between efficiency and expressive capacity.

\section{Method}
\label{sec:method}

\textbf{SRLoRA} addresses the expressiveness bottleneck of fixed-rank LoRA by dynamically recomposing the update subspace over the course of training. 
Unlike existing approaches that statically allocate a fixed subspace, SRLoRA identifies under-utilized LoRA components, fuses them into the base model, and reinitializes their subspace using unused singular directions from the pretrained weight matrix. 
This mechanism enables LoRA to explore a larger functional subspace without increasing the total number of trainable parameters.

\begin{figure}[tbp]
\centering
\begin{tabular}[t]{cc}
\subfigure[SVD on pretrained weights]{\label{subfig-svd}\includegraphics[width=0.495\textwidth]{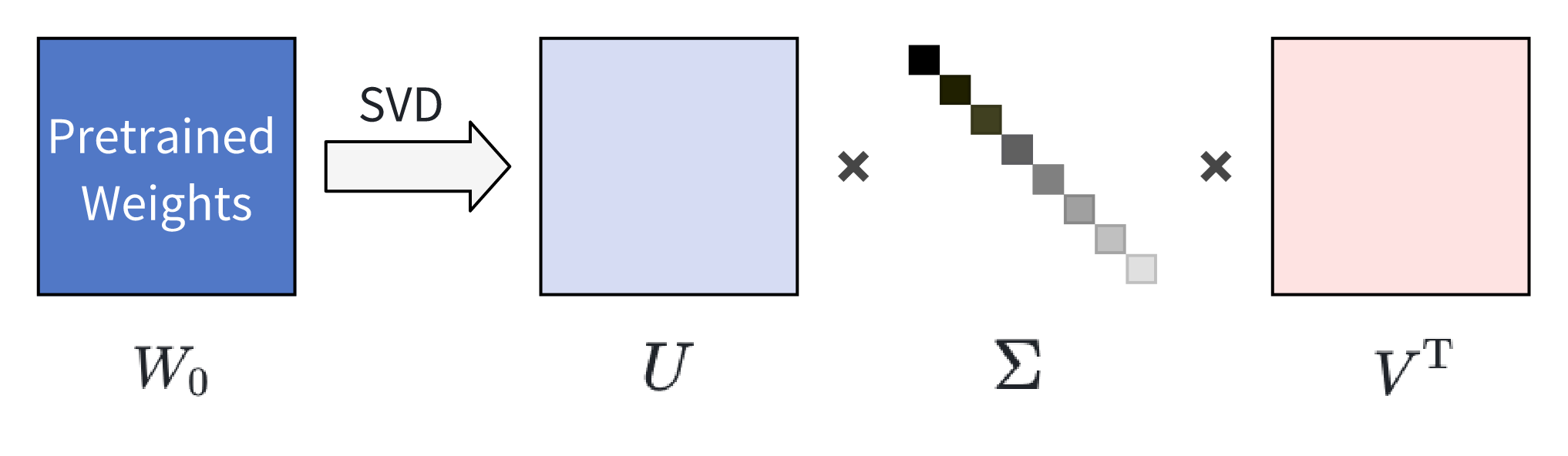}}
\subfigure[Fusing low-importance pairs]{\label{subfig-imp}\includegraphics[width=0.495\textwidth]{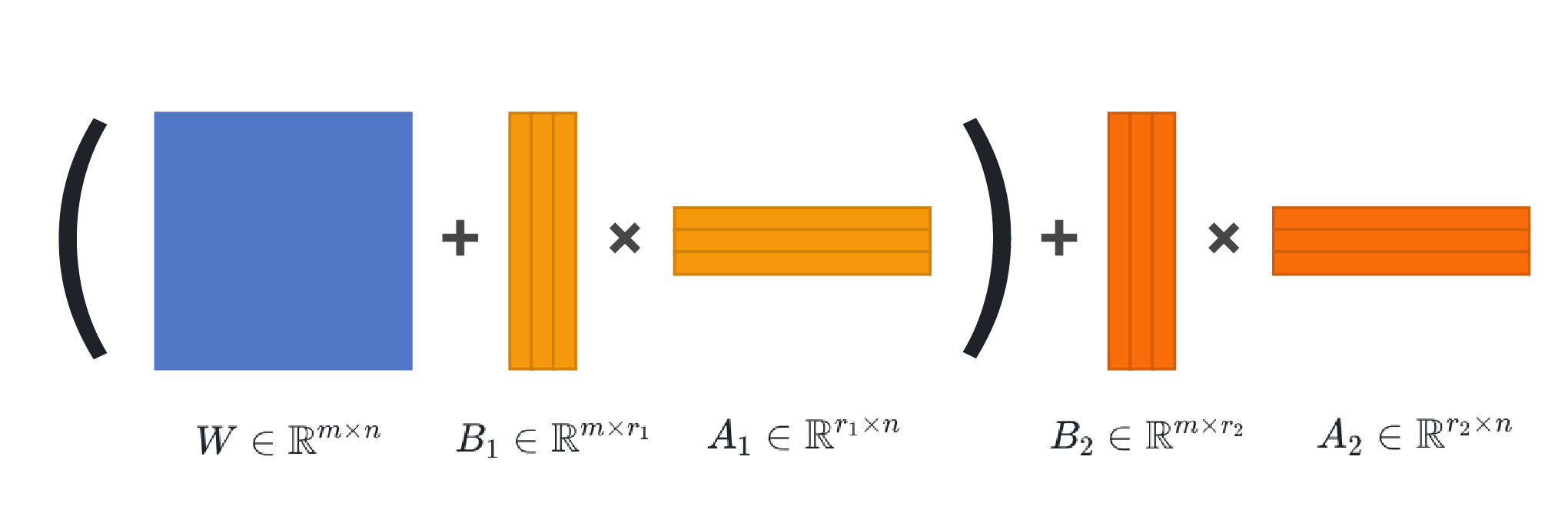}}
\end{tabular}
\caption{Overview of SRLoRA. (a) SVD of pretrained weights. (b) Fusion of low-importance LoRA components into the base model. Lighter orange areas indicate lower importance scores, representing components that are being fused into the base model.}
\label{fig:svd-imp}
\end{figure}

The method proceeds in four key stages: 
(i) perform SVD on the frozen pretrained weights to extract a rank-ordered basis (see Fig. \ref{subfig-svd}); 
(ii) estimate pairwise importance scores for each rank-1 LoRA component using a sensitivity-based criterion; 
(iii) fuse low-importance components into the base model and discard their trainable parameters (see Fig. \ref{subfig-imp}); and 
(iv) reinitialize these ranks using the next unused SVD directions, while subtracting their contributions from the base weights to prevent redundancy. We now present our proposed method.


\textbf{SVD of pretrained weights.} Given a frozen pretrained weight matrix $\mW_0 \in \mbr{m \times n}$, we compute its SVD:
\begin{equation}
\mW_0 = \mU \Sigma \mV^\top,
\end{equation}
where $\mU \in \mbr{m \times d}$ and $\mV \in \mbr{n \times d}$ are orthonormal matrices containing the left and right singular vectors, respectively, and $\Sigma = \mathrm{diag}(\sigma_1, \ldots, \sigma_d)$ contains the singular values in descending order. Here, $d = \min(m, n)$.
LoRA introduces a trainable update $\Delta \mW = \mB\mA$ of rank $r \ll d$, where $\mB \in \mbr{m \times r}$ and $\mA \in \mbr{r \times n}$. We adopt the PiSSA~\cite{meng2024pissa} initialization, which aligns the adaptation subspace with the top-$r$ singular directions of $W_0$:
\begin{equation}
\mB = \mU_{[:, :r]} \Sigma_{[:r, :r]}^{1/2}, \quad \mA = \Sigma_{[:r, :r]}^{1/2} \mV_{[:, :r]}^\top.
\end{equation}

This initialization ensures that the low-rank projection $\Delta \mW$ starts aligned with the most important axes in the pretrained parameter space, promoting rapid convergence and stability.

\subsection{Importance-Based Fusion and Reinitialization}

\textbf{Sensitivity-based importance estimation.} The sensitivity of model parameters is commonly estimated using the gradient-weight product, a metric originally proposed in \cite{michel2019sixteenheadsreallybetter} and later refined  in \cite{liang2021superticketspretrainedlanguage} and \cite{zhang2022platonpruninglargetransformer}. This score quantifies how much the training loss is affected by each individual parameter in the model. The sensitivity for a parameter at position $(i, j)$ is defined as:
\begin{equation}
I(w_{ij}) = \left| w_{ij} \cdot \nabla_{w_{ij}} \mathcal{L} \right|,
\label{eq:importance}
\end{equation}
where $w_{ij}$ is the value of the model parameter and $\nabla_{w_{ij}} \mathcal{L}$ denotes the gradient of the loss function $\mathcal{L}$ with respect to that parameter. The absolute value ensures the importance is \textit{non-negative}, regardless of gradient direction.
This formulation offers a first-order Taylor approximation of the increase in loss when the parameter $w_{ij}$ is zeroed out. Intuitively, if removing a parameter leads to a large increase in loss, the model is said to be highly sensitive to it. Therefore, $I(w_{ij})$ serves as a proxy for parameter importance\cite{michel2019sixteenheadsreallybetter, liang2021superticketspretrainedlanguage, zhang2022platonpruninglargetransformer}.

However, this sensitivity measure can exhibit high variance when computed over a single mini-batch. This is due to the inherent randomness in batch sampling and the stochastic nature of training dynamics. As a result, relying on raw sensitivity scores can lead to noisy or unstable importance estimates \cite{zhang2022platonpruninglargetransformer}.
To mitigate this issue, Zhang \etal\cite{zhang2022platonpruninglargetransformer} introduced a smoothing technique that stabilizes importance estimation through exponential moving averages (EMA) of both the sensitivity and its uncertainty.
%
The smoothed sensitivity of parameter $w_{ij}$ at iteration $t$ is defined as:
\begin{equation}
\bar{I}^{(t)}(w_{ij}) = \beta_1 \bar{I}^{(t-1)}(w_{ij}) + (1 - \beta_1) I^{(t)}(w_{ij}),
\end{equation}
where $\beta_1 \in (0, 1)$ is a smoothing coefficient that balances the contribution of past and current importance values. Here, $I^{(t)}(w_{ij})$ denotes the instantaneous importance of $w_{ij}$ at iteration $t$, while $\bar{I}^{(t)}(w_{ij})$ represents its exponentially smoothed estimate. This formulation aggregates historical importance signals to yield a more stable and robust measure over time.


In addition to tracking smoothed sensitivity, we also estimate the local uncertainty of each parameter's importance using a separate EMA:
\begin{equation}
\bar{U}^{(t)}(w_{ij}) = \beta_2 \bar{U}^{(t-1)}(w_{ij}) + (1 - \beta_2) \left| I^{(t)}(w_{ij}) - \bar{I}^{(t)}(w_{ij}) \right|,
\end{equation}
where $\beta_2 \in (0, 1)$ controls the rate of adaptation to recent changes. This uncertainty estimate captures the short-term variability of the importance score, measuring how consistently a parameter's contribution fluctuates around its smoothed value $\bar{I}^{(t)}(w_{ij})$.

To compute the final importance score for parameter $w_{ij}$ at iteration $t$, we take the product of its smoothed sensitivity and its estimated uncertainty:
\begin{equation}
s^{(t)}(w_{ij}) = \bar{I}^{(t)}(w_{ij}) \cdot \bar{U}^{(t)}(w_{ij}).
\end{equation}

This formulation encourages the model to prioritize parameters that are not only consistently important (reflected in a high $\bar{I}$), but also stable and reliable (indicated by a low deviation captured by $\bar{U}$).
While AdaLoRA\cite{zhang2023adaptive} computes importance over SVD triplets, our method retains the original LoRA decomposition where the low-rank update is modeled as $\Delta \mW \!= \!\mB\mA$, with $\mB \!\in\! \mbr{m \times r}$ and $\mA \!\in \!\mbr{r \times n}$. Instead of evaluating triplets, we assess the importance of individual rank-1 components derived from each column-row pair in $\mB$ and $\mA$.
For the $k$-th rank-1 component, the importance score is computed by aggregating the importance scores of the corresponding column in $\mB$ and row in $\mA$:
\begin{equation}
S_k^{(t)} = \frac{1}{m} \sum_{i=1}^{m} s^{(t)}(\mB_{ik}) + \frac{1}{n} \sum_{j=1}^{n} s^{(t)}(\mA_{kj}),
\end{equation}
where $\mB_{ik}$ is the $i$-th element of the $k$-th column in $\mB$, and $\mA_{kj}$ is the $j$-th element of the $k$-th row in $\mA$. This score provides a robust and interpretable metric for identifying low-importance LoRA components that can later be fused or reinitialized.

\textbf{Fusing low-importance pairs.} At predefined intervals, we rank the LoRA components by their importance scores $\{S_k^{(t)}\}$ and identify the lowest-ranked subset. We define fusion ratio $\gamma$ and let $r'$ denote the number of ranks to recycle, where $r'=\gamma\cdot r$. We split the update as:
\begin{equation}
\Delta \mW = \sum_{k=1}^r \mB_{\cdot k} \mA_{k \cdot} = \underbrace{\sum_{k \in \mathcal{I}_\text{low}} \mB_{\cdot k} \mA_{k \cdot}}_{\text{fused into } \mW} + \underbrace{\sum_{k \in \mathcal{I}_\text{high}} \mB_{\cdot k} \mA_{k \cdot}}_{\text{retained for training}},
\end{equation}
where $\mathcal{I}_\text{low}$ and $\mathcal{I}_\text{high}$ denote the index sets of low- and high-importance pairs, respectively.

The low-importance update is fused into the frozen weight matrix:
\begin{equation}
\mW \leftarrow \mW + \sum_{k \in \mathcal{I}_\text{low}} \mB_{\cdot k} \mA_{k \cdot},
\end{equation}
after which the corresponding LoRA parameters $\mB_{\cdot k}$ and $\mA_{k \cdot}$ are discarded, reducing the active parameter count.








\textbf{Reinitialization of LoRA components.}
After identifying and fusing low-importance LoRA components, we reinitialize the corresponding low-rank matrices to explore new subspaces not previously used. This reinitialization is guided by the SVD of the frozen pretrained weight matrix $\mW_0$.
At the beginning of training, we initialize the LoRA matrices $\mA$ and $\mB$ using the PiSSA method, which selects the top-$r$ principal components of $\mW_0$ to provide an informed starting subspace.
During the subspace recomposition phase, we identify low-importance component pairs $\mB_1$ $\mA_1$ and fuse their contributions into the current frozen weights $\mW$:
\begin{equation}
    \mW \leftarrow \mW + \mB_1 \mA_1.
\end{equation}

We then reinitialize these ranks using the next unused singular directions from the SVD of $\mW_0$. Specifically, if $\mU \mS \mV^\top = \mathrm{SVD}(\mW_0)$, and $p_r$ denotes the index of the last used singular direction, we select the next $r'$ singular vectors to construct new bases:
\begin{align}
\left\{
\begin{aligned}
& \mB_1^{\mathrm{new}} = \mU_{[:, p_r:p_r+r']}\mS_{[p_r:p_r+r', p_r:p_r+r']}^{1/2} \\
& \mA_1^{\mathrm{new}} = \mS_{[p_r:p_r+r', p_r:p_r+r']}^{1/2}\mV_{[:, p_r:p_r+r']}^\top
\end{aligned}
\right.
\end{align}
where $r'$ is the number of ranks being reinitialized.

To avoid duplicating subspace contributions, we subtract the newly initialized low-rank projection from the current frozen weight matrix:
\begin{equation}
    \mW \leftarrow \mW - \mB_1^{\mathrm{new}} \mA_1^{\mathrm{new}}.
\end{equation}

This subtraction ensures that the newly added directions are orthogonal to the currently active subspace, preserving the model’s representation diversity. Finally, we reset the importance scores of all LoRA components to prepare for the next training interval.

\begin{algorithm}[tbp]
\caption{SRLoRA: Subspace Recomposition for Efficient LoRA Fine-tuning}
\label{alg:srlora}
\begin{algorithmic}[1]
\STATE \textbf{Input:} Dataset $\mathcal{D}$; pretrained weights $\mW_0$; training iterations $T$; switching schedule $\mathcal{T}_{\text{switch}}$; smoothing factors $\beta_1, \beta_2$; fusion ratio $\gamma$.
\STATE \textbf{Initialize:} LoRA matrices $\mA$, $\mB$ via PiSSA using top-$r$ singular vectors of $\mW_0$; set importance scores $s^{(0)} = 0$.
\FOR{$t = 1$ to $T$}
    \IF{$t \in \mathcal{T}_{\text{switch}}$}
        \STATE Compute component-wise importance scores $S_k^{(t)}$.
        \STATE Select lowest $\gamma$ fraction of components based on $S_k^{(t)}$.
        \STATE Fuse selected low-importance components $\mB_1 \mA_1$ into current frozen matrix $\mW$: $\mW \leftarrow \mW + \mB_1 \mA_1$
        \STATE Reinitialize $\mB_1, \mA_1$ using next unused singular directions:
        $\mB_1^{\mathrm{new}} = \mU_{[:, p_r:p_r+r']} \Sigma^{1/2}, \quad
        \mA_1^{\mathrm{new}} = \Sigma^{1/2} \mV_{[:, p_r:p_r+r']}^\top$
        \STATE Subtract new projection to avoid duplication:
        $\mW \leftarrow \mW - \mB_1^{\mathrm{new}} \mA_1^{\mathrm{new}}$
        \STATE Reset importance scores for reinitialized ranks.
    \ELSE
        \STATE Sample mini-batch from $\mathcal{D}$ and compute gradients $\nabla \mathcal{L}(\mW)$.
        \STATE Compute sensitivity:
        $I^{(t)}(w_{ij}) = \left| w_{ij} \cdot \nabla_{w_{ij}} \mathcal{L} \right|$
        \STATE Update smoothed sensitivity:
        $\bar{I}^{(t)}(w_{ij}) = \beta_1 \bar{I}^{(t-1)}(w_{ij}) + (1 - \beta_1) I^{(t)}(w_{ij})$
        \STATE Update uncertainty:
        $\bar{U}^{(t)}(w_{ij}) = \beta_2 \bar{U}^{(t-1)}(w_{ij}) + (1 - \beta_2) \left| I^{(t)}(w_{ij}) - \bar{I}^{(t)}(w_{ij}) \right|$
        \STATE Compute final importance:
        $s^{(t)}(w_{ij}) = \bar{I}^{(t)}(w_{ij}) \cdot \bar{U}^{(t)}(w_{ij})$
        \STATE Aggregate importance for LoRA component $k$:
        $S_k^{(t)} = \frac{1}{m} \sum_{i=1}^m s^{(t)}(\mB_{ik}) + \frac{1}{n} \sum_{j=1}^n s^{(t)}(\mA_{kj})$
        \STATE Update trainable LoRA parameters $\mB$, $\mA$ via gradient descent.
    \ENDIF
\ENDFOR
\STATE \textbf{Output:} Fine-tuned model parameters $\mW^{(T)}$.
\end{algorithmic}
\end{algorithm}

\subsection{Subspace-Recomposed LoRA}

We now introduce \textbf{SRLoRA}, a dynamic low-rank adaptation strategy that iteratively recomposes low-rank subspace by identifying and fusing unimportant LoRA components into pretrained model weights, and reinitializing them using previously unused singular directions. This allows the model to progressively explore new, orthogonal low-rank subspaces that better capture task-specific information. 
SRLoRA is detailed in Algorithm~\ref{alg:srlora}.

To quantify the extent to which SRLoRA can expand its representational subspace, we define a hyperparameter $r_{\text{target}}$, representing the maximum allowed subspace rank that SRLoRA can explore. Based on this, the number of switch operations is computed as:
\begin{equation}
N_{\text{switch}} =  \frac{r_{\text{target}} - r}{r'}
\end{equation}
where $r$ is the initial LoRA rank and $r'$ is the number of rank to recycle. Consequently, the switching interval $t_{\text{interval}}$ is defined as:
\begin{equation}
t_{\text{interval}} =  \frac{N_{\text{all}}}{N_{\text{switch}}},
\end{equation}
where $N_{\text{all}}$ denotes the total number of training steps, and $N_{\text{switch}}$ is the total number of switching events.
%
 At regular intervals specified by the switch iteration set $\mathcal{T}_{\text{switch}} = \{ t_{\text{interval}},\\\, 2t_{\text{interval}},\, 3t_{\text{interval}}, \, \ldots \}$, SRLoRA computes the importance scores of LoRA components and selectively fuses the least important ones into the frozen pretrained weights. 
The freed-up ranks are then reinitialized using unused singular vectors from the SVD of the original pretrained weights. To maintain orthogonality and prevent duplication, the newly initialized components are subtracted from the frozen weights.
Between switching intervals, standard training proceeds with gradient-based updates, during which sensitivity-based importance scores are tracked using exponential moving averages. These scores guide the next fusion and reinitialization cycle. 

In the next section, we present our experiments and results.

\begin{table}[tbp]
\begin{center}
\resizebox{0.8\linewidth}{!}{\begin{tabular}{lcl}
\toprule
\textbf{Task Name} & \textbf{Metric} & \textbf{Task Description} \\
\midrule
SST-2  & Accuracy         & The Stanford Sentiment Treebank \\
MRPC   & F1 / Accuracy    & Microsoft Research Paraphrase Corpus \\
CoLA   & Matthew's Corr   & The Corpus of Linguistic Acceptability \\
QNLI   & Accuracy         & Question Natural Language Inference \\
RTE    & Accuracy         & Recognizing Textual Entailment \\
STS-B  & Pearson-Spearman Corr   & Semantic Textual Similarity Benchmark \\
\toprule
\end{tabular}}
\end{center}
\caption{Tasks used in GLUE with corresponding evaluation metrics and descriptions.}
\label{tab:glue_tasks}
\end{table}

\section{Experiment}
\label{sec:exp}

\subsection{Setup}

\textbf{GLUE benchmark.} We evaluate on six representative GLUE tasks covering single-sentence classification (SST-2 for sentiment, CoLA for grammatical acceptability), sentence-pair classification (MRPC for paraphrase detection, QNLI for question answering, RTE for textual entailment), and sentence-pair regression (STS-B for semantic similarity). Details are summarized in Table~\ref{tab:glue_tasks}.


\textbf{Image benchmark.} Our method is also tested on CIFAR-100, STL-10, and MNIST, spanning diverse resolutions and complexities. CIFAR-100 has 60,000 color images across 100 classes; STL-10 offers 5,000 labeled images over 10 classes plus 100,000 unlabeled samples; MNIST contains 70,000 grayscale handwritten digit images. These datasets provide a broad evaluation of model adaptability.

\begin{table}[tbp]
  \centering
  \resizebox{\textwidth}{!}{\begin{tabular}{lcccccccccccc}
  \toprule
    \multirow{2}{*}{\textbf{Dataset}}
      & \multicolumn{3}{c}{\textbf{LoRA}}
      &&  \multicolumn{3}{c}{\textbf{PiSSA}}
      && \multicolumn{4}{c}{\textbf{SRLoRA}} \\
    \cline{2-4}
    \cline{6-8}
    \cline{10-13}
      & Epoch & BS & LR&
      & Epoch & BS & LR&
      & Epoch & BS & LR & Target Rank \\
   \midrule
    SST-2  & 20 & 16 & 3e-5 && 20 & 16 & 3e-5 && 20 & 16 & 3e-5 & 512 \\
    MRPC   & 20 & 32 & 2e-4 && 20 & 32 & 2e-4 && 20 & 32 & 2e-4 & 256\\
    CoLA   & 20 & 16 & 1e-4 && 20 & 16 & 1e-4 && 20 & 16 & 1e-4 & 256\\
    QNLI   & 10 & 32 & 1e-4 && 10 & 32 & 1e-4 && 10 & 32 & 1e-4 & 128\\
    RTE    & 50 & 16 & 1e-4 && 50 & 16 & 1e-4 && 50 & 16 & 1e-4 & 16\\
    STS-B  & 20 &  8 & 1e-4 && 20 &  8 & 1e-4 && 20 &  8 & 1e-4 & 32\\
  \bottomrule
  \end{tabular}}
  \caption{Hyperparameters used for LoRA, PiSSA and SRLoRA on DeBERTa-v3-base. BS and LR refer to the batch size and learning rate, respectively.}
\label{tab:hyper}
\end{table}

\begin{figure}[tbp]
\centering
\begin{tabular}[t]{cc}
\subfigure[Training loss curves on the RTE task.]{\label{fig:loss}\includegraphics[width=0.49\textwidth]{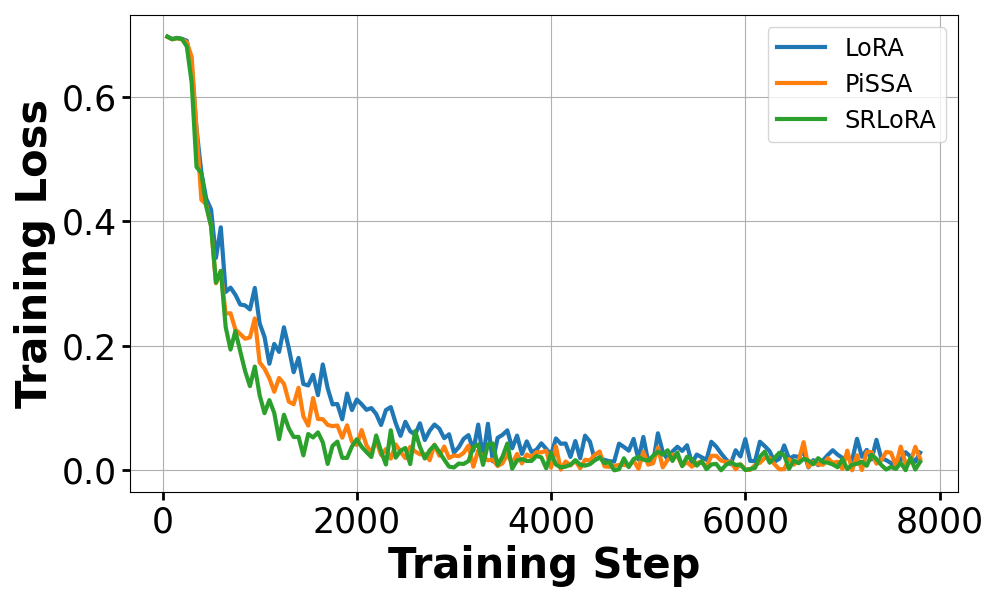}} &
\subfigure[Training loss curves on the QNLI task.]{\label{fig:loss2}\includegraphics[width=0.49\textwidth]{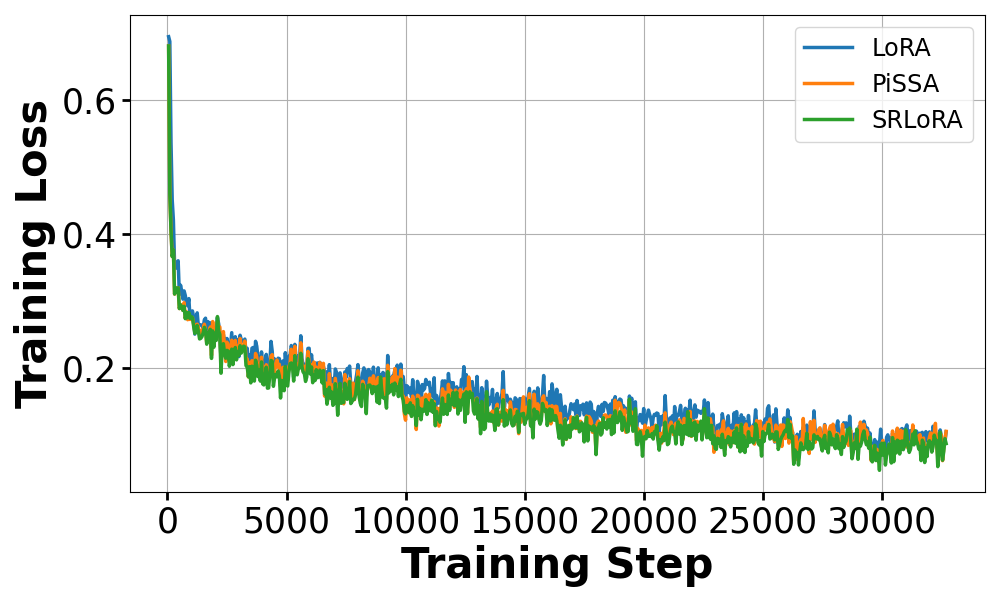}} \\

\subfigure[Training loss curves on the SST2 task.]{\label{fig:acc1}\includegraphics[width=0.49\textwidth]{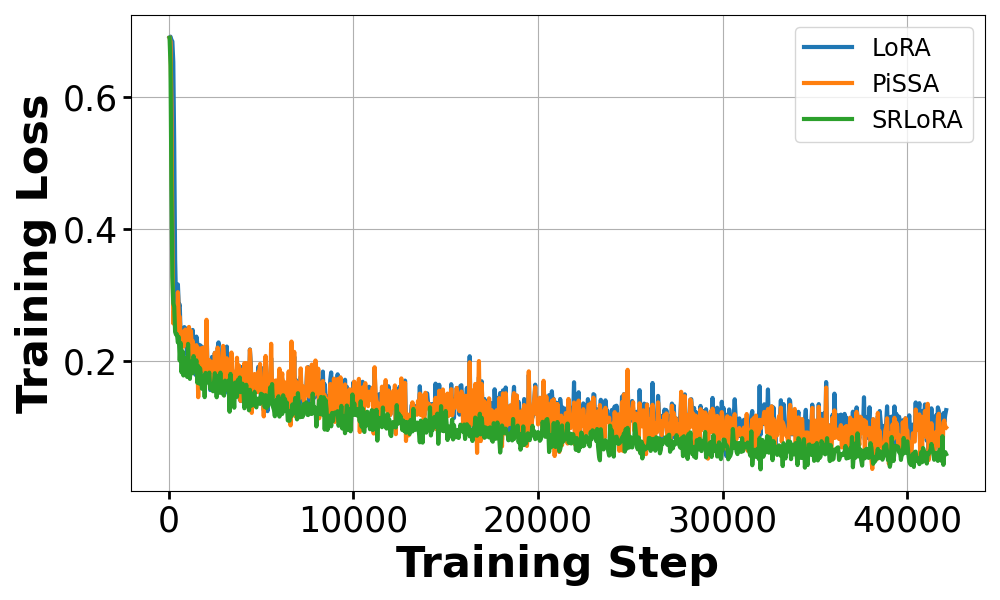}} &
\subfigure[Training loss curves on the CoLA task.]{\label{fig:acc2}\includegraphics[width=0.49\textwidth]{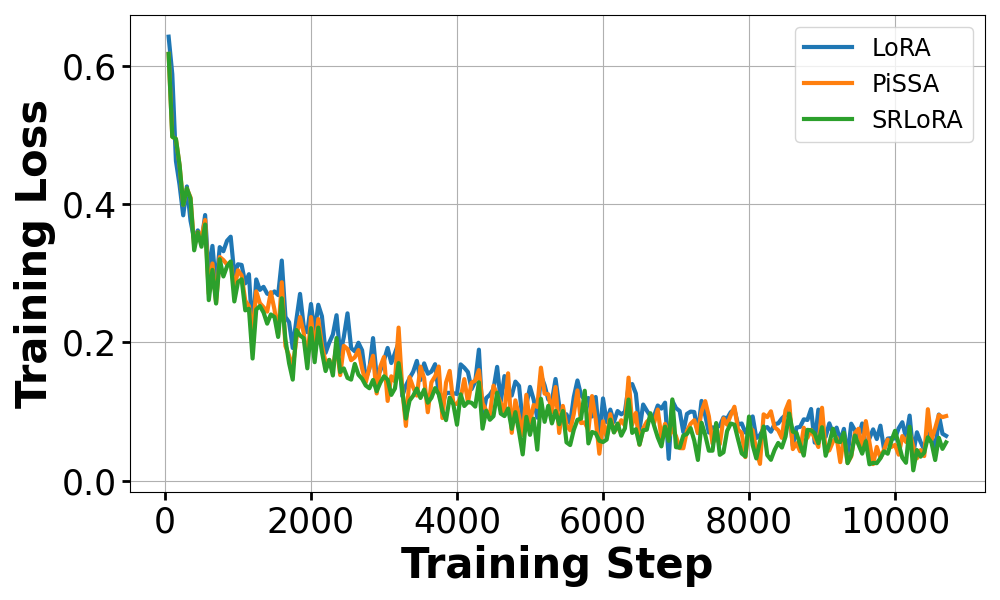}}

\end{tabular}
\caption{Training loss comparison on RTE, QNLI, SST2 and CoLA tasks. SRLoRA demonstrates faster convergence and improved stability over baseline methods.}
\label{fig:loss-comp}
\end{figure}

\textbf{Implementation details.}
We compare SRLoRA with LoRA and PiSSA on a subset of the GLUE benchmark using the DeBERTa-v3-base model~\cite{he2023debertav}. Experiments run on a single AMD Instinct MI250X GPU with PyTorch (FP32) and the PEFT library. AdamW is used consistently across methods. Hyperparameters, including learning rate, batch size, and epochs, are listed in Table~\ref{tab:hyper}.

For all GLUE tasks, the maximum sequence length is fixed at 128 tokens, with a low-rank dimension $r=8$, scaling factor $\alpha=8$, and no dropout to ensure fair comparison. SRLoRA performs subspace fusion at each switching step with a fusion ratio $\gamma=0.5$. We adopt default settings $\beta_1 = \beta_2 = 0.85$ from \cite{zhang2023adaptive}, set warm-up steps to 500, and apply no weight decay. Models are evaluated every 500 steps, selecting the checkpoint with the lowest validation loss as final.
To validate generalizability, we also evaluate on image classification using Vision Transformers. We use ViT-B/16 pretrained on ImageNet-21K (vit-b16-224-in21k), training with PyTorch Lightning 2.0.2 and 16-bit mixed precision on a single GPU. LoRA rank and scaling factor are fixed at $r=8$ and $\alpha=8$. Training runs for 5000 steps with an SGD optimizer, cosine learning rate scheduler, and a 500-step warm-up. Evaluation is conducted every 500 steps.

\begin{figure}
\centering
\begin{tabular}{c}
\bmvaHangBox{\fbox{\includegraphics[width=0.95\textwidth]{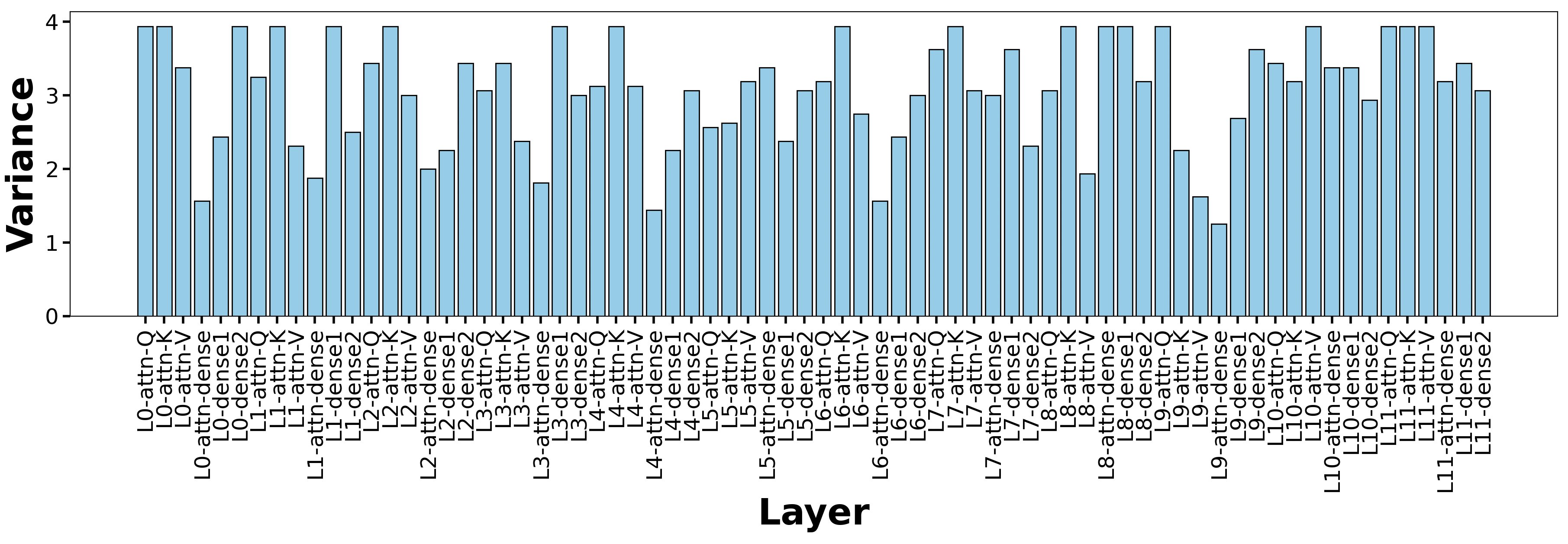}}}\\
\end{tabular}
\caption{
Variance of activation intervals across different candidate ranks for each SRLoRA-enabled layer in DeBERTa-v3-base on the CoLA task. ``Active intervals'' refer to the rank positions that receive significant updates during low-rank adaptation. Lower variance in the Feedforward (FFN) layers suggests that many rank positions are equally important, indicating a need to update a broader subspace. In contrast, attention projection layers (Q, K, V) show high variance, meaning only a few rank directions dominate, aligned with the presence of large singular values.
}
\label{fig:variance}
\end{figure}

\begin{figure}[tbp]
\centering
\begin{tabular}{c}
\bmvaHangBox{\fbox{\includegraphics[width=0.95\textwidth]{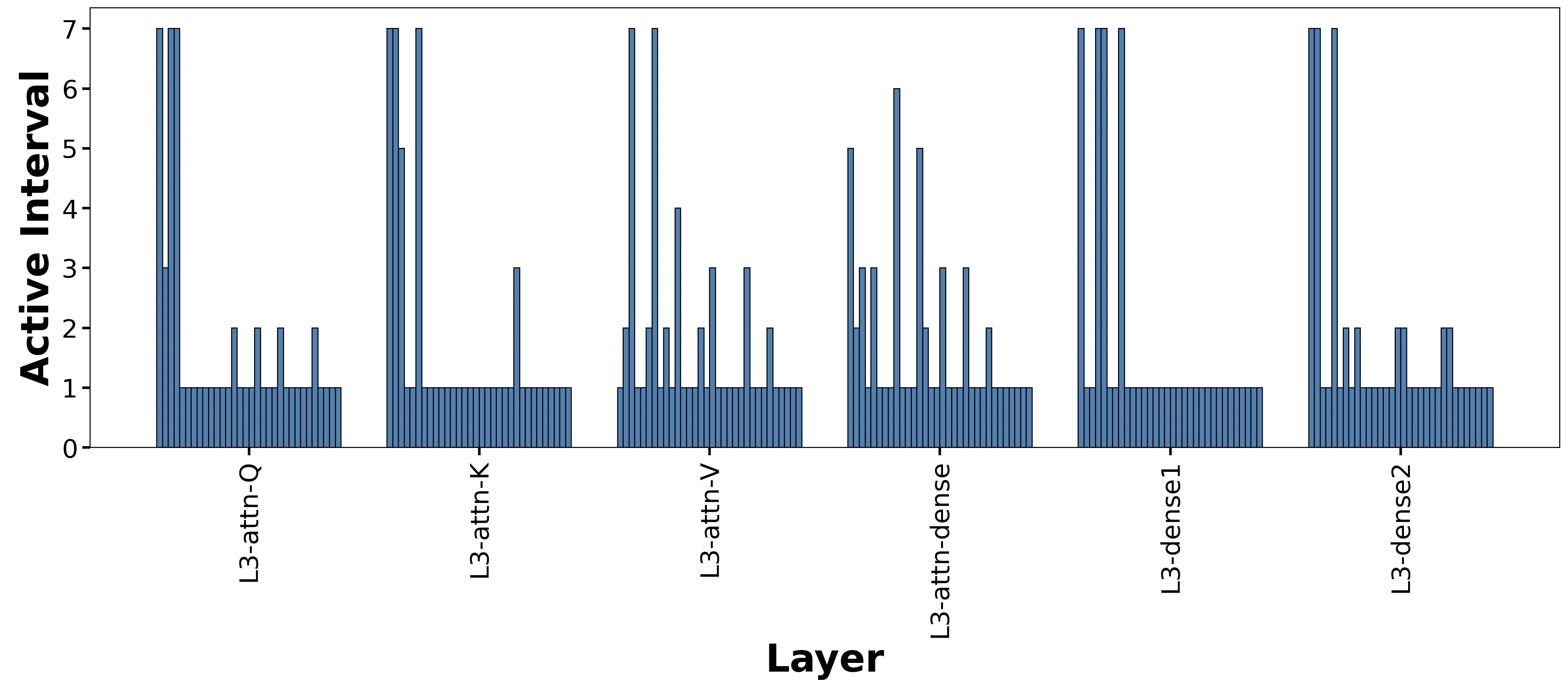}}}\\
\end{tabular}
\caption{Active intervals for target ranks in Layer 3 of DeBERTa-v3-base on the CoLA task. Each interval represents the duration (in training steps or epochs) that a given SRLoRA rank remains actively updated before reverting to the frozen base weights. Ranks corresponding to larger singular values (shown on the left) tend to stay active longer, indicating their greater importance in adaptation. In contrast, dense (FFN) layers exhibit a more evenly distributed activation pattern, with ranks switching more frequently and balanced across the subspace.
}
\label{fig:iterval}
\end{figure}

\begin{table}[tbp]
\begin{center}
\resizebox{\linewidth}{!}{\begin{tabular}{lccccccc}
\toprule
Method & Params/Total Params & SST2 & MRPC & CoLA & QNLI & RTE & STSB \\
\midrule
LoRA    & 1.33M/184M & 95.9 & \textbf{90.8/87.5} & \textbf{65.4} & 94.0 & 80.4 & \textbf{90.5}/89.8 \\
PiSSA   & 1.33M/184M & 95.7 & 90.5/87.2 & 64.7 & \textbf{94.4} & 82.0 & 89.6/89.0 \\
SRLoRA  & 1.33M/184M & \textbf{96.1} & 90.3/86.6 & 65.1 & 93.4 & \textbf{82.1} & 90.4/\textbf{90.2} \\
\bottomrule
\end{tabular}}
\end{center}
\caption{Comparison of LoRA, PiSSA and SRLoRA on GLUE benchmarks. Bold numbers indicate the best performance in each task.}
\end{table}

\subsection{Evaluation}

\textbf{Faster and more effective early training with SRLoRA.}
Figure~\ref{fig:loss} shows the training loss curves when fine-tuning DeBERTa-v3-base using LoRA, PiSSA, and our proposed SRLoRA under identical hyperparameters. Notably, SRLoRA achieves a significantly faster reduction in training loss during the initial phase, converging to a lower loss within the first 2,000 steps compared to LoRA and PiSSA. 
This accelerated convergence reflects SRLoRA’s superior ability to exploit the low-rank subspace early in training through its subspace recomposition strategy, which dynamically fuses low-importance components back into pretrained weights and reinitializes them with unused principal directions. This continual refreshment of the low-rank representation enables more expressive subspaces and efficient gradient updates, driving faster optimization.

\textbf{Dominance of high-rank singular vectors in subspace adaptation.}
Figure~\ref{fig:variance} quantifies the variance of active intervals for SRLoRA components across layers, derived from detailed per-pair analysis shown in Figure~\ref{fig:iterval}. 
Each SRLoRA-enhanced module maintains 32 candidate rank pairs corresponding to SVD-initialized singular vectors ordered by descending singular values. 
We observe that pairs associated with larger singular values consistently remain active for longer intervals during training, aligning with the importance scoring defined in Eq.~\eqref{eq:importance}. 
A high variance in active intervals indicates that top-ranked singular vectors dominate adaptation, resulting in a narrower subspace focus. Conversely, lower variance reflects more uniform participation across ranks, suggesting greater subspace flexibility and adaptation capacity at higher ranks. This insight shows how SRLoRA selectively emphasizes critical principal components while dynamically managing representational capacity.

\textbf{Robust performance gains on challenging vision tasks.}
Extending evaluation beyond NLP, we compare LoRA and SRLoRA on ViT models fine-tuned on CIFAR-100, STL10, and MNIST datasets (Table \ref{tab:img}). 
SRLoRA demonstrates clear advantages in complex tasks like CIFAR-100, where the ability to dynamically reallocate representational power is crucial for improved performance. 
While traditional LoRA remains competitive on simpler or saturated tasks such as MNIST, SRLoRA’s adaptive reinitialization and fusion mechanisms enable it to achieve lower training loss and improved convergence stability on more demanding datasets. 

Notably, after 4,000 training steps, while the loss values across methods converge, SRLoRA consistently maintains a slight advantage, highlighting its improved training stability and efficiency enabled by continuous subspace recomposition.


\begin{table}[tbp]
\centering
\begin{tabular}{lccccccc}
\toprule
\textbf{Method} & \textbf{CIFAR-100} & \textbf{STL10} & \textbf{MNIST} \\
\midrule
LoRA    & 90.06 & \textbf{99.62} & \textbf{98.89} \\
SRLoRA  & \textbf{92.51} & 99.54 & 94.83 \\
\bottomrule
\end{tabular}
\caption{LoRA \textit{vs.} SRLoRA on ViT fine-tuned for CIFAR-100, STL-10, and MNIST.}
\label{tab:img}
\end{table}

\section{Conclusion}
\label{sec:conclusion}

We propose SRLoRA, a novel method that dynamically enhances low-rank adaptation by fusing unimportant LoRA components back into frozen weights and reinitializing them using unused principal directions from the pretrained weight’s SVD. Unlike static approaches like PiSSA, SRLoRA recycles underutilized subspace dimensions during training, enabling continuous adaptation without increasing trainable parameters.
Our experiments on GLUE benchmarks and Vision Transformer image classification tasks demonstrate SRLoRA’s effectiveness, especially on complex tasks where static low-rank methods fall short. While gains are less pronounced on simpler datasets, SRLoRA offers a general and efficient way to boost LoRA fine-tuning adaptability.
Future work includes developing an adaptive switch scheduling strategy that triggers fusion and reinitialization based on training dynamics, such as importance score convergence or loss stagnation, to ensure optimal training of each subspace component and enhance overall efficiency and stability.

\section*{Acknowledgments}
Haodong Yang conducted this research under the supervision of Lei Wang and Md Zakir Hossain as part of his final year master's research project at ANU.
This work was supported by computational resources provided by the Pawsey Supercomputing Centre, a high-performance computing facility funded by the Australian Government.




\bibliography{egbib}








\end{document}